\documentclass[letterpaper]{article}

\usepackage{aaai}
\usepackage{times}
\usepackage{helvet}
\usepackage{courier}
\usepackage{algorithmic}

\nocopyright

\usepackage[utf8]{inputenc}
\usepackage{natbib}
\usepackage[british]{babel}
\usepackage{relsize}
\usepackage{stmaryrd}
\usepackage{url}
\usepackage{color}
\usepackage{multicol}
\usepackage[font=it]{caption}
\usepackage{tikz,pgf}
\usetikzlibrary{shapes,backgrounds}
\usepackage{amsmath,amssymb}
\usepackage[amsmath,thref,thmmarks]{ntheorem}
\usepackage{cleveref}
\usepackage{multirow}
\usepackage{xspace}
\usepackage{comment}
\usepackage{setspace}
\usepackage{paralist}
\usepackage{algorithm}
\usepackage{float}
\usepackage{fancyvrb} 
\usepackage{lastpage} %
\usepackage{ifthen}
\usepackage{environ}
\newif\iflong
\NewEnviron{shortproof}{\iflong\else\expandafter\begin{proof}\BODY\end{proof}\fi}
\NewEnviron{longproof}{\iflong\expandafter\begin{proof}\BODY\end{proof}\fi}
\longtrue
\longfalse

\theoremstyle{plain}
\theoremseparator{.}
\theoremheaderfont{\normalfont\bfseries}
\theorembodyfont{\slshape}
\newtheorem{theorem}{Theorem}

\newtheorem{proposition}[theorem]{Proposition}
\newtheorem{corollary}[theorem]{Corollary}
\theorembodyfont{\upshape}
\theoremsymbol{\ensuremath{\blacktriangle}}
\newtheorem{definition}{Definition}
\theoremsymbol{\ensuremath{\blacksquare}}
\newtheorem{example}{Example}
\theoremstyle{nonumberplain}
\theoremheaderfont{\normalfont\itshape}

\theoremstyle{nonumberplain}
\theoremheaderfont{\normalfont\itshape}
\theoremsymbol{} %
\newtheorem{proof}{Proof}
\newcommand{\mysubsubsection}[1]{\vspace*{-1.1em}\paragraph{#1}}

\ifdefined\implies
 \renewcommand{\implies}{\rightarrow}
\else
\newcommand{\implies}{\rightarrow}
\fi
\newcommand{\limplies}{\rightarrow}
\newcommand{\liff}{\leftrightarrow}

\newcommand{\define}[1]{\emph{#1}}
\newcommand{\set}[1]{\left\{#1\right\}}
\newcommand{\guard}{\ \middle\vert\ }

\newcommand{\abs}[1]{\left\lvert#1\right\rvert}

\newcommand{\phiff}{\mathrel{\phantom{\iff}}}
\newcommand{\update}[3]{#1|^{#2}_{#3}}

\newcommand{\tvt}{\ensuremath{\mathbf{t}}\xspace}
\newcommand{\tvf}{\ensuremath{\mathbf{f}}\hspace*{0.005mm}\xspace}
\newcommand{\tvu}{\ensuremath{\mathbf{u}}\xspace}
\newcommand{\tvx}{\ensuremath{\mathbf{x}}\xspace}

\newcommand{\two}{\ensuremath{\set{\tvt,\tvf}}\xspace}
\newcommand{\three}{\ensuremath{\set{\tvt,\tvf,\tvu}}\xspace}

\renewcommand{\int}[1]{#1}
\newcommand{\allv}{\ensuremath{\mathcal{V}}\xspace}
\newcommand{\tve}[1]{\ensuremath{[{#1}]_2}\xspace}
\newcommand{\topv}[1]{\ensuremath{\max_{\ileq}#1}\xspace}
\newcommand{\topvt}[1]{\ensuremath{\max_{\tleq}#1}\xspace}

\newcommand{\adm}{\mathit{adm}} %
\newcommand{\com}{\mathit{com}} %
\newcommand{\pre}{\mathit{prf}} %
\newcommand{\prf}{\mathit{prf}} %
\renewcommand{\mod}{\mathit{mod}} %

\newcommand{\adf}{\ensuremath{D}\xspace}
\newcommand{\parents}[1]{\ensuremath{\mathit{par}(#1)}\xspace}
\newcommand{\voc}{\ensuremath{A}\xspace}
\newcommand{\ileq}{\ensuremath{\leq_i}\xspace}
\newcommand{\ilt}{\ensuremath{<_i}\xspace}

\newcommand{\tleq}{\ensuremath{\leq_t}\xspace}
\newcommand{\tlt}{\ensuremath{<_t}\xspace}

\newcommand{\expressive}{e}

\newcommand{\eleq}{\leq_\expressive}

\newcommand{\lt}{<}
\newcommand{\elt}{\lt_\expressive}

\newcommand{\F}{\mathcal{F}}
\newcommand{\kb}{\mathsf{kb}}
\newcommand{\AF}{\textrm{AF}}
\newcommand{\SETAF}{\textrm{SETAF}}
\newcommand{\BADF}{\textrm{BADF}}
\newcommand{\ADF}{\textrm{ADF}}

\newcommand{\signature}[2]{\ensuremath{\Sigma_{#1}^{#2}}}
\newcommand{\copSym}[1]{\ensuremath{\Gamma_{#1}}\xspace}
\newcommand{\cop}[2]{\ensuremath{\copSym{#1}({#2})}\xspace}

\newcommand{\realize}{\ensuremath{\mathit{realize}}}
\newcommand{\realizePrf}{\ensuremath{\mathit{realizePrf}}}

\newcommand{\propagator}[2]{\ensuremath{\mathit{P}^{#1}_{#2}}\xspace}
\newcommand{\canonKB}[2]{\ensuremath{\mathit{kb}^{#1}_{#2}}\xspace}

 \newcommand{\tool}[1]{\texttt{#1}\xspace}
\newcommand{\gringo}{\tool{Gringo}}
\newcommand{\aspartix}{\tool{ASPARTIX}}
\newcommand{\diamondT}{\tool{DIAMOND}}

\longtrue

\title{Characterizing Realizability in \mbox{Abstract Argumentation}\thanks{This research has been supported by DFG (project BR 1817/7-1) and FWF (projects I1102 and P25518).}}
\author{Thomas Linsbichler\\
TU Wien\\
Austria\\
\And
Jörg Pührer \and Hannes Strass\\
Leipzig University\\
Germany}

\begin{document}
\maketitle

\begin{abstract}
  Realizability for knowledge representation formalisms studies the following question:
  Given a semantics and a set of interpretations, is there a knowledge base whose semantics coincides exactly with the given interpretation set?
  We introduce a general framework for analyzing realizability in abstract dialectical frameworks (ADFs) and various of its subclasses.
  In particular, the framework applies to Dung argumentation frameworks, SETAFs by Nielsen and Parsons, and bipolar ADFs.
  We present a uniform characterization method for the admissible, complete, preferred and model/stable semantics.
  We employ this method to devise an algorithm that decides realizability for the mentioned formalisms and semantics;
  moreover the algorithm allows for constructing a desired knowledge base whenever one exists.
  The algorithm is built in a modular way and thus easily extensible to new formalisms and semantics.
  We have also implemented our approach in answer set programming, and used the implementation to obtain several novel results on the relative expressiveness of the abovementioned formalisms.
\end{abstract}

\section{Introduction}

The abstract argumentation frameworks (AFs) introduced by \citet{Dung1995} have garnered increasing attention in the recent past.
In his seminal paper, \citeauthor{Dung1995} showed how an abstract notion of argument (seen as an atomic entity) and the notion of individual attacks between arguments together could reconstruct several established KR formalisms in argumentative terms.
Despite the generality of those and many more results in the field that was sparked by that paper, researchers also noticed that the restriction to \emph{individual attacks} is often overly limiting, and devised extensions and generalizations of Dung's frameworks\/:
directions included generalizing individual attacks to \emph{collective attacks}~\citep{NielsenP06}, leading to so-called \SETAF{}s;
others started offering a \emph{support} relation between arguments~\citep{DBLP:conf/ecsqaru/CayrolL05a},
preferences among arguments~\citep{DBLP:journals/amai/AmgoudC02,DBLP:journals/ai/Modgil09},
or attacks on attacks into arbitrary depth \citep{DBLP:journals/ijar/BaroniCGG11}.
This is only the tip of an iceberg, for \iflong{a more comprehensive overview we refer to the work of }\fi\citet{BrewkaPW2014}.

One of the most recent and most comprehensive generalizations of AFs has been presented by \citet{BrewkaW2010} (and later continued by \citealp{BrewkaESWW2013}) in the form of \emph{abstract dialectical frameworks (ADFs)}.
These ADFs offer any type of link between arguments:
individual attacks (as in AFs), collective attacks (as in \SETAF{}s), and individual and collective support, to name only a few.
This generality is achieved through so-called \emph{acceptance conditions} associated to each statement.
Roughly, the meaning of relationships between arguments is not fixed in ADFs, but is specified by the user for each argument in the form of Boolean functions (acceptance functions) on the argument's parents.
\iflong{%
However, this generality comes with a price:
\citet{StrassW2015} found that the complexity of the associated reasoning problems of ADFs is in general higher than in AFs (one level up in the polynomial hierarchy).
Fortunately, the subclass of \emph{bipolar ADFs} (defined by \citealp{BrewkaW2010}) is as complex as AFs (for all considered semantics) while still offering a wide range of modeling capacities~\citep{StrassW2015}.
However, there has only been little concerted effort so far to exactly analyze and compare the expressiveness of the abovementioned languages.
}\fi

This paper is about exactly analyzing means of expression for argumentation formalisms.
\iflong{Instead of motivating expressiveness in natural language and showing examples that some formalisms seem to be able to express but others do not, we tackle the problem in a formal way. }\fi
We use a precise mathematical definition of expressiveness:
a set of interpretations is \emph{realizable} by a formalism under a semantics if and only if there exists a knowledge base of the formalism whose semantics is exactly the given set of interpretations.
Studying realizability in AFs has been started by \iflong{\citet*{DunneDLW2013,DunneDLW2015}}\fi, who analyzed realizability for extension-based semantics, that is, interpretations represented by sets where arguments are either accepted (in the extension set) or not accepted (not in the extension set).
While their initial work disregarded arguments that are never accepted, there have been continuations where the existence of such ``invisible'' arguments is ruled out \citep{BaumannDLSW2014,LinsbichlerSW15}.
\citet{Dyrkolbotn2014} began to analyze realizability for labeling-based semantics of AFs, that is, three-valued semantics where arguments can be accepted (mapped to true), rejected (mapped to false) or neither (mapped to unknown).
\citet{Strass2015} started to analyze the relative expressiveness of two-valued semantics for ADFs (relative with respect to related formalisms).
\iflong{
Most recently, \citet{Puehrer2015} presented precise characterizations of realizability for ADFs under several three-valued semantics, namely admissible, grounded, complete, and preferred.
The term ``precise characterizations'' means that he gave necessary and sufficient conditions for an interpretation~set to be ADF-realizable under a semantics.
}\fi

\iflong{%
The present paper continues this line of work by lifting it to a much more general setting.
  We combine the works of \citet{DunneDLW2015}, \citet{Puehrer2015}, and \citet{Strass2015} into a unifying framework, and at the same time extend them to formalisms and semantics not considered in the respective papers:%
}\fi\ 
we treat several formalisms, namely AFs, \SETAF{}s, and (B)ADFs, while the previous works all used different approaches and techniques.
This is possible because all of these formalisms can be seen as subclasses of ADFs that are obtained by suitably restricting the acceptance conditions.

Another important feature of our framework is that we uniformly use three-valued interpretations as the underlying model theory.
In particular, this means that arguments cannot be ``invisible'' any more since the underlying vocabulary of arguments is always implicit in each interpretation.
Technically, we always assume a fixed underlying vocabulary and consider our results parametric in that vocabulary.
In contrast, for example, \citet{Dyrkolbotn2014} presents a construction for realizability that introduces new arguments into the realizing knowledge base;
we do not allow that.
While sometimes the introduction of new arguments can make sense, for example if new information becomes available about a domain or a debate,
it is not sensible in general, as these new arguments would be purely technical with an unclear dialectical meaning.
Moreover, it would lead to a different notion of realizability, where most of the realizability problems would be significantly easier, if not trivial.

The paper proceeds as follows.
\iflong{%
We begin with recalling and introducing the basis and basics of our work -- the formalisms we analyze and the methodology with which we analyze them.
}\fi
Next we introduce our general framework for realizability;
the major novelty is our consistent use of so-called characterization functions, firstly introduced by \citet{Puehrer2015}, which we adapt to further semantics.
The main workhorse of our approach will be a parametric propagate-and-guess algorithm for deciding whether a given interpretation set is realizable in a formalism under a semantics.
We then analyze the relative expressiveness of the considered formalisms, presenting several new results that we obtained using an implementation of our framework.
We conclude with a discussion.

\section{Preliminaries}

We make use of standard mathematical concepts like functions and partially ordered sets.
For a function \mbox{$f:X\to Y$} we denote the \define{update of $f$ with a pair \mbox{$(x,y)\in X\times Y$}} by \mbox{$\update{f}{x}{y}:X\to Y$} with \mbox{$z\mapsto y$} if \mbox{$z=x$}, and \mbox{$z\mapsto f(z)$} otherwise.
For a function \mbox{$f:X\to Y$} and \mbox{$y\in Y$}, its preimage is \mbox{$f^{-1}(y)=\set{ x\in X \guard f(x)=y }$}.
A \define{partially ordered set} is a pair $(S,\sqsubseteq)$ with $\sqsubseteq$ a partial order on $S$. %
A partially ordered set $(S,\sqsubseteq)$ is a \define{complete lattice} if and only if every \mbox{$S'\subseteq S$} has both a greatest lower bound (glb) \mbox{$\bigsqcap S'\in S$} and a least upper bound (lub) \mbox{$\bigsqcup S'\in S$}.
A partially ordered set $(S,\sqsubseteq)$ is a \define{complete meet-semilattice} iff every non-empty subset \mbox{$S'\subseteq S$} has a greatest lower bound \mbox{$\bigsqcap S'\in S$} (the \define{meet}) and every ascending chain \mbox{$C\subseteq S$} has a least upper bound \mbox{$\bigsqcup C\in S$}.

\mysubsubsection{Three-Valued Interpretations}
Let $\voc$ be a fixed finite set of statements.
An \define{interpretation} is a mapping \mbox{$v:\voc\to\three$} that assigns one of the truth values true ($\tvt$), false ($\tvf$) or unknown ($\tvu$) to each statement.
An interpretation is \define{two-valued} if \mbox{$v(\voc)\subseteq\two$}, that is, the truth value $\tvu$ is not assigned.
Two-valued interpretations $v$ can be extended to assign truth values \mbox{$v(\varphi)\in\two$} to propositional formulas $\varphi$ as usual.

The three truth values are partially ordered according to their information content:
we have \mbox{$\tvu\ilt\tvt$} and \mbox{$\tvu\ilt\tvf$} and no other pair in $\ilt$, which intuitively means that the classical truth values contain more information than the truth value unknown.
As usual, we denote by $\ileq$ the partial order associated to the strict partial order $\ilt$.
The pair $(\set{\tvt,\tvf,\tvu},\ileq)$ forms a complete meet-semilattice with the information meet operation $\sqcap_i$.
This meet can intuitively be interpreted as \define{consensus} and assigns
\mbox{$\tvt\sqcap_i\tvt = \tvt$}, \mbox{$\tvf\sqcap_i\tvf = \tvf$}, and returns $\tvu$ otherwise.

The information ordering $\ileq$ extends in a straightforward way to interpretations $v_1,v_2$ over $\voc$ in that
\mbox{$v_1 \ileq v_2$} iff \mbox{$v_1(a) \ileq v_2(a)$} for all \mbox{$a\in \voc$}.
We say for two interpretations $v_1,v_2$ that $v_2$ \define{extends} $v_1$ iff \mbox{$v_1 \ileq v_2$}.
The set \allv of all interpretations over $\voc$ forms a complete meet-semilattice with respect to the information ordering $\ileq$.
The consensus meet operation $\sqcap_i$ of this semilattice is given by \mbox{$(v_1\sqcap_i v_2)(a)= v_1(a)\sqcap_i v_2(a)$} for all \mbox{$a\in \voc$}.
The least element of $(\allv,\ileq)$ is the valuation \mbox{$v_\tvu:\voc\to\set{\tvu}$} mapping all statements to unknown -- the least informative interpretation.
\iflong{By $\allv_2$ we denote the set of two-valued interpretations;
they are the $\ileq$-maximal elements of the meet-semilattice $(\allv,\ileq)$.
We denote by $\tve{v}$ the set of all two-valued interpretations that extend~$v$.
The elements of $\tve{v}$ form an $\ileq$-antichain with greatest lower bound \mbox{$v=\bigsqcap_i\tve{v}$}.%
}\fi
\mysubsubsection{Abstract Argumentation Formalisms}
An \emph{abstract dialectical framework (ADF)} is a tuple \mbox{$D = (\voc, L, C)$} where
$\voc$ is a set of statements (representing positions one can take or not take in a debate),
\mbox{$L \subseteq \voc\times\voc$} is a set of links (representing dependencies between the positions),
\mbox{$C = \set{C_a}_{a \in \voc}$} is a collection of functions 
\mbox{$C_a : 2^{\parents{a}}\to \set{\tvt,\tvf}$}, one for each statement \mbox{$a\in\voc$}. 
The function $C_a$ is the \define{acceptance condition of $a$} and expresses whether $a$ can be accepted, given the acceptance status of its parents \mbox{$\parents{a}=\set{b\in S\guard (b,a)\in L}$}.
We usually represent each $C_a$ by a propositional formula \mbox{$\varphi_a$} over $\parents{a}$.
To specify an acceptance condition, then, we take \mbox{$C_a(M\cap\parents{a})=\tvt$} to hold iff $M$ is a model for $\varphi_a$.

\citet{BrewkaW2010} introduced a useful subclass of ADFs:
an ADF \mbox{$D = (\voc, L, C)$} is \define{bipolar} iff all links in $L$ are supporting or attacking (or both).
A link \mbox{$(b,a) \in L$} is \define{supporting in $D$} iff for all \mbox{$M \subseteq \parents{a}$}, we have that \mbox{$C_a(M)=\tvt$} implies \mbox{$C_a(M \cup \set{b})=\tvt$}.
Symmetrically, a link \mbox{$(b,a) \in L$} is \define{attacking in $D$} iff for all \mbox{$M \subseteq \parents{a}$}, we have that \mbox{$C_a(M \cup \set{b})=\tvt$} implies \mbox{$C_a(M)=\tvt$}.
If a link \mbox{$(b,a)$} is both supporting and attacking then $b$ has no actual influence on $a$.
(But the link does not violate bipolarity.)
We write BADFs as \mbox{$D=(\voc, L^+\cup L^-, C)$} and mean that $L^+$ contains all supporting links and $L^-$ all attacking links.

\iflong{%
The semantics of ADFs can be defined using an operator $\copSym{D}$ over three-valued interpretations~\citep{BrewkaW2010,BrewkaESWW2013}.
For an ADF~$D$ and a three-valued interpretation~$v$, the interpretation $\cop{D}{v}$ is given by
\begin{gather*}
  a \mapsto {\textstyle\bigsqcap_i} \set{ w(\varphi_a) \guard w\in\tve{v} }
\end{gather*}
That is, for each statement $a$, the operator returns the consensus truth value for its acceptance formula $\varphi_a$, where the consensus takes into account all possible two-valued interpretations $w$ that extend the input valuation $v$.
If this $v$ is two-valued, we get \mbox{$\tve{v}=\set{v}$} and thus \mbox{$\cop{D}{v}(a) = v(\varphi_a)$}.
}\fi

The standard semantics of ADFs are now defined as follows.
For ADF $D$, an interpretation \mbox{$v:\voc\to\three$} is
\iflong{%
  \begin{itemize}
  \item \define{admissible} iff \mbox{$v\ileq\cop{D}{v}$};
  \item \define{complete} iff \mbox{$\cop{D}{v} = v$};
  \item \define{preferred} iff it is $\ileq$-maximal admissible;
  \item a \define{two-valued model} iff it is two-valued and \mbox{$\cop{D}{v}=v$}.
  \end{itemize}
}\fi
We denote the sets of interpretations that are
admissible, complete, preferred, and two-valued models %
by $\adm(D)$, $\com(D)$, $\prf(D)$ and $\mod(D)$, respectively.
These definitions are proper generalizations of Dung's notions for AFs: %
For an AF \mbox{$(A, R)$},
where $R \subseteq A \times A$ is the attack relation,
the \define{ADF associated to $(A,R)$} is
\mbox{$D_{(A,R)}=(A,R,C)$} with \mbox{$C=\{\varphi_a\}_{a\in A}$} and
\mbox{$\varphi_a=\bigwedge_{b:(b,a)\in R}\neg b$} for all \mbox{$a\in A$}.
AFs inherit their semantics from the definitions for ADFs~\citep[Theorems~2~and~4]{BrewkaESWW2013}.
In particular, an interpretation is \define{stable} for an AF $(A,R)$ if and only if it is a two-valued model of $D_{(A,R)}$.
{\fussy
A \SETAF{} is a pair \mbox{$S = (\voc,X)$} %
where
\mbox{$X \subseteq (2^\voc \setminus \{\emptyset\}) \times \voc$}
is the (set) attack relation.
We define three-valued counterparts of
the semantics introduced by \citet{NielsenP06},
following the same conventions as in
three-valued semantics of AFs \citep{CaminadaG09} and
argumentation formalisms in general.
Given a statement \mbox{$a \in \voc$} and an interpretation $v$
we say that
$a$ is \define{acceptable} wrt.\ $v$
if \mbox{$\forall (B,a) \in X \exists a' \in B : v(a') = \tvf$} and
$a$ is \define{unacceptable} wrt.\ $v$
if \mbox{$\exists (B,a) \in X \forall a' \in B : v(a') = \tvt$}.
For an interpretation \mbox{$v:\voc\to\three$} it holds that
\iflong{
\begin{itemize}
\item $v \in \adm(S)$ iff for all $a \in \voc$,
							 $a$ is acceptable wrt.\ $v$ if $v(a) = \tvt$ and
							 $a$ is unacceptable wrt.\ $v$ if $v(a) = \tvf$;
\item $v \in \com(S)$ iff for all $a \in \voc$,
							$a$ is acceptable wrt.\ $v$ iff $v(a) = \tvt$ and
							$a$ is unacceptable wrt.\ $v$ iff $v(a) = \tvf$;
\item $v \in \prf(S)$ iff $v$ is $\ileq$-maximal admissible; and
\item $v \in \mod(S)$ iff $v\in\adm(F)$ and $\nexists a \in \voc : v(a) = \tvu$.
\end{itemize}
For a \SETAF{} \mbox{$S=(\voc,X)$}
the corresponding ADF $D_S$
has acceptance formula \mbox{$\varphi_a = \bigwedge_{(B,a) \in X} \bigvee_{a' \in B} \neg a'$} for each statement \mbox{$a\in\voc$}.~\citep{PolbergPhD2016}
\begin{proposition}
\label{prop:setaf_to_adf}
For any \SETAF{} \mbox{$S=(\voc,X)$} it holds that \sloppy
\mbox{$\sigma(S) = \sigma(D_S)$},
where \mbox{$\sigma \in \{\adm,\com,\prf,\mod\}$}.
\begin{longproof}
Given interpretation $v$ and statement $a$, it holds that
\mbox{$\cop{D_S}{v}(a) = \tvt$} iff \mbox{$\forall w \in \tve{v} : w(a) = \tvt$} iff
\sloppy \mbox{$\forall (B,a) \in X$} \mbox{$\exists a' \in B : v(a') = \tvf$} iff $a$ is acceptable wrt.\ $v$ and
\mbox{$\cop{D_S}{v}(a) = \tvf$} iff \mbox{$\forall w \in \tve{v} : w(a) = \tvf$} iff
\mbox{$\exists (B,a) \in X$} \mbox{$\forall a' \in B : v(a') = \tvt$} iff $a$ is unacceptable wrt.\ $v$.
Hence \mbox{$\sigma(S) = \sigma(D_S)$} for \mbox{$\sigma \in \{\adm,\com,\prf,\mod\}$}. \hfill$\Box$
\end{longproof}
\end{proposition}
}\fi
}%
\mysubsubsection{Realizability}
A set \mbox{$V\subseteq\allv$} of interpretations is \define{realizable} in a formalism $\F$ under a semantics $\sigma$
if and only if there exists a knowledge base \mbox{$\kb \in \F$} having exactly \mbox{$\sigma(\kb) = V$}.
\citet{Puehrer2015} characterized realizability for ADFs under various three-valued semantics.
We will reuse the central notions for capturing the complete semantics in this work.
\begin{definition}[\citealt{Puehrer2015}]\label{def:comChar}
Let $V$ be a set of interpretations.
A function \mbox{$f:\allv_2 \to \allv_2$} is a \define{$\com$-characterization} of $V$ iff: for each \mbox{$v\in \allv$} we have \mbox{$v\in V$} iff for each \mbox{$a\in \voc$}\/:
\begin{itemize}
\item \mbox{$v(a)\neq \tvu$} implies \mbox{$f(v_2)(a)=v(a)$} for all \mbox{$v_2\in\tve{v}$} and
\item \mbox{$v(a)= \tvu$} implies \mbox{$f(v'_2)(a)=\tvt$} and \mbox{$f(v''_2)(a)=\tvf$} for some \mbox{$v'_2,v''_2\in\tve{v}$}.
\end{itemize}
\end{definition}
From a function of this kind we can build a corresponding ADF by the following construction.
For a function \mbox{$f:\allv_2 \to \allv_2$}, we define $\adf_{f}$ as the ADF where the acceptance formula for each statement $a$ is given by
\begin{gather*}
  \varphi^f_a=\mathop{\bigvee_{w\in \allv_2,}}_{f(w)(a)=\tvt} \phi_w
  \quad\text{with}\quad
  \phi_w = \bigwedge_{w(a')=\tvt} a' \land \bigwedge_{w(a')=\tvf} \neg a'
\end{gather*}
Observe that we have \mbox{$v(\phi_w)=\tvt$} iff \mbox{$v=w$} by definition.
Intuitively, the acceptance condition $\varphi_a^f$ is constructed such that $v$ is a model of $\varphi_a^f$ if and only if we find \mbox{$f(v)(a)=\tvt$}.

\begin{proposition}[\citealt{Puehrer2015}]\label{prop:comReal}
  Let \mbox{$V\subseteq\allv$} be a set of interpretations.
  (1)~For each ADF \adf with \mbox{$\com(\adf)=V$}, there is a $\com$-characterization $f_\adf$ for $V$;
  (2)~for each $\com$-char\-acterization \mbox{$f:\allv_2\to\allv_2$} for $V$ we have \mbox{$\com(\adf_f)=V$}.
\end{proposition}%
\iflong{The result shows that $V$ can be realized under complete semantics if and only if there is a $\com$-characterization for $V$.}\fi

\section{A General Framework for Realizability}
\label{sec:general-framework}

The main underlying idea of our framework is that all abstract argumentation formalisms introduced in the previous section can be viewed as subclasses of \iflong{abstract dialectical frameworks}\fi.
This is clear for ADFs themselves and for BADFs by definition;
for AFs and \SETAF{}s it is fairly easy to see.
However, knowing that these formalisms can be recast as ADFs is not everything.
To employ this knowledge for realizability, we must be able to precisely characterize the corresponding subclasses in terms of restricting the ADFs' acceptance functions.
Alas, this is also possible and paves the way for the framework we present in this section.
Most importantly, we will make use of the fact that different formalisms and different semantics can be characterized modularly, that is, independently of each other.

Towards a uniform account of realizability for ADFs under different semantics,
we start with a new characterization of realizability for ADFs under admissible semantics
that is based on a notion similar in spirit to $\com$-characterizations.
\begin{definition}\label{def:admChar}
Let $V$ be a set of interpretations.
A function $f:\allv_2 \to \allv_2$ is an \define{$\adm$-characterization} of $V$ iff:
for each $v\in \allv$ we have $v\in V$ iff for every $a\in \voc$\/:
\begin{itemize}
\item $v(a)\neq \tvu$ implies $f(v_2)(a)=v(a)$ for all $v_2\in\tve{v}$.
\end{itemize}
\end{definition}
Note that the only difference to \Cref{def:comChar} is dropping the second condition related to statements with truth value \tvu.

\begin{proposition}\label{prop:admReal}
  Let \mbox{$V\subseteq\allv$} be a set of interpretations.
  (1)~For each ADF \adf such that \mbox{$\adm(\adf)=V$}, there is an $\adm$-characterization $f_\adf$ for $V$;
  (2)~for each adm-cha\-racterization \mbox{$f:\allv_2\to\allv_2$} for $V$ we have \mbox{$\adm(\adf_f)=V$}.
\begin{longproof}
(1) %
    We define the function $f_\adf:\allv_2 \to \allv_2$ as 
    $f_\adf(v_2)(a)=v_2(\varphi_a)$
    for every $v_2\in \allv_2$ and $a\in \voc$ where $\varphi_a$ is the acceptance formula of $a$ in $\adf$.
    We will show that $f_\adf$ is an $\adm$-characterization for $V=\adm(\adf)$.
    Let $v$ be an interpretation. Consider the case $v\in \adm(\adf)$
    and $v(a)\neq u$ for some $a\in \voc$ and some $v_2\in\tve{v}$.
    From $v\ileq\cop{\adf}{v}$ we get $v_2(\varphi_a)=v(a)$.
    By definition of $f_\adf$ is follows that $f_\adf(v_2)(a)=v(a)$.
    Now assume $v\not\in \adm(\adf)$ and consequently $v\not\ileq\cop{\adf}{v}$.
    There must be some $a\in \voc$ such that $v(a)\neq\tvu$ and $v(a)\neq \cop{\adf}{v}(a)$.
    Hence, there is some $v_2\in\tve{v}$ with $v_2(\varphi_a)\neq v(a)$
    and $f_\adf(v_2)(a)\neq v(a)$ by definition of $f_\adf$. 
    Thus, $f_\adf$ is an $\adm$-characterization %
    
(2)
    Observe that for every two-valued interpretation $v_2$
    and every $a\in \voc$ we have
    $f(v_2)(a)=v_2(\varphi^f_a)$.
$(\subseteq)$:
      Let $v\in\adm(\adf_{f})$ be an interpretation and $a\in \voc$ a statement
      such that $v(a)\neq\tvu$.
      Let $v_2$ be a two-valued interpretation with $v_2\in\tve{v}$.
      Since $v\ileq\cop{\adf_{f}}{v}$ we have $v(a)=v_2(\varphi^f_a)$.
      Therefore, by our observation it must also hold that $f(v_2)(a)=v(a)$.
      Thus, by Definition~\ref{def:admChar}, %
      $v\in V$.
$(\supseteq)$:
      Consider an interpretation $v$ such that $v\not\in\adm(\adf_{f})$.
      We show that $v\not\in V$.
      From $v\not\in\adm(\adf_{f})$ we get $v\not\ileq\cop{\adf_{f}}{v}$.
      There must be some $a\in \voc$ such that $v(a)\neq\tvu$ and $v(a)\neq \cop{\adf_{f}}{v}(a)$.
      Hence, there is some $v_2\in\tve{v}$ with $v_2(\varphi^f_a)\neq v(a)$
      and consequently $f(v_2)(a)\neq v(a)$. Thus, by Definition~\ref{def:admChar} we have $v\not\in V$.
      \hfill$\Box$
  \end{longproof}
\end{proposition}
When listing sets of interpretations in examples, for the sake of readability we represent three-valued interpretations by sequences of truth values, tacitly assuming that the underlying vocabulary is given and has an associated total ordering.
For example, for the vocabulary \mbox{$\voc=\set{a,b,c}$} we represent the interpretation \mbox{$\set{ a\mapsto\tvt,b\mapsto\tvf,c\mapsto\tvu }$} by the sequence~$\tvt\tvf\tvu$.
\begin{example}\label{ex:admChar}
Consider the sets \mbox{$V_1=\{\tvu\tvu\tvu,\tvt\tvf\tvf,\tvf\tvt\tvu\}$} and \mbox{$V_2=\{\tvt\tvf\tvf,\tvf\tvt\tvu\}$} of interpretations over \mbox{$\voc=\set{a,b,c}$}.
The mapping
$f=\{ 
\tvt\tvt\tvt \mapsto \tvf\tvt\tvt,
\tvt\tvt\tvf \mapsto \tvt\tvf\tvt,
\tvt\tvf\tvt \mapsto \tvt\tvt\tvt,
\tvt\tvf\tvf \mapsto \tvt\tvf\tvf,
\tvf\tvt\tvt \mapsto \tvf\tvt\tvf,
\tvf\tvt\tvf \mapsto \tvf\tvt\tvt,
\tvf\tvf\tvt \mapsto \tvt\tvt\tvf,
\tvf\tvf\tvf \mapsto \tvf\tvt\tvf
\}$ is an $\adm$-characterization for $V_1$.
Thus, the ADF $\adf_{f}$ has $V_1$ as its admissible interpretations.
\iflong{Indeed, the realizing ADF has the following acceptance conditions\/:
$$
\begin{array}{lll}
\varphi^f_a & \equiv & (a \land b \land \neg c) \lor (a \land \neg b) \lor (\neg a \land \neg b \land c)\\
\varphi^f_b & \equiv & (a \land c) \lor (\neg a \land b) \lor (\neg a \land \neg b \land \neg c)\\
\varphi^f_c & \equiv & (a \land b) \lor (\neg a \land b \land \neg c) \lor (\neg b \land c)
\end{array}
$$
}\fi
For $V_2$ no $\adm$-characterization exists because \mbox{$\tvu\tvu\tvu\not\in V_2$} but the implication of \Cref{def:admChar} trivially holds for $a$, $b$, and~$c$.
\end{example}

We have seen that the construction $\adf_{f}$ for realizing under complete semantics can also be used for realizing a set $V$ of interpretations under admissible semantics.
\iflong{The only difference is that we here 
}\fi 
require $f$ to be an $\adm$-characterization instead of a $\com$-characterization for $V$.
Note that admissible semantics can be characterized by properties that are easier to check than existence of an $\adm$-characterization \iflong{\citep[see the work of][]{Puehrer2015}}\fi. 
However, using the same type of characterizations for different semantics allows for a unified approach for checking realizability and constructing a realizing ADF in case one exists.

For realizing under the model semantics, we can likewise present an adjusted version of $\com$-characterizations.
\begin{definition}
  \label{def:modChar}
  Let \mbox{$V\subseteq\allv$} be a set of interpretations.
  A function \mbox{$f:\allv_2 \to \allv_2$} is a \define{$\mod$-characterization} of $V$ if and only if:
  (1) $f$ is defined on $V$ (that is, \mbox{$V\subseteq\allv_2$}) and 
  (2) for each \mbox{$v\in\allv_2$}, we have \mbox{$v\in V$} iff \mbox{$f(v)=v$}.
\end{definition} 

As we can show, there is a one-to-one correspondence between $\mod$-characterizations and ADF realizations.
\begin{proposition}
  \label{prop:modReal}
  Let \mbox{$V\subseteq\allv$} be a set of interpretations.
  (1)~For each ADF \adf such that \mbox{$\mod(\adf)=V$}, there is a $\mod$-characterization $f_\adf$ for $V$;
  (2)~vice versa, for each $\mod$-characterization \mbox{$f:\allv_2\to\allv_2$} for $V$ we find \mbox{$\mod(\adf_f)=V$}.%
  \begin{longproof}
    (1) Let \adf be an ADF with \mbox{$\mod(\adf)=V$}.
    It immediately follows that \mbox{$V\subseteq\allv_2$}.
    To define $f_\adf$ we can use the construction in the proof of \Cref{prop:admReal}.
    It follows directly that for any \mbox{$v\in\allv_2$}, we find \mbox{$f_D(v)=v$} iff \mbox{$v\in V$}.
    Thus $f_\adf$ is a $\mod$-characterization for $V$.
    
    (2) Let \mbox{$V\subseteq\allv_2$} and \mbox{$f:\allv_2\to\allv_2$} be a $\mod$-characterization of $V$.
    For any \mbox{$v\in\allv_2$} we have\/:
    \begin{align*}
      &\phiff v\in V
      \iff v=f(v) \\
      &\iff \forall a\in\voc: \left( v(a)=f(v)(a) \right) \\
      &\iff \forall a\in\voc: \left( v(a)=\tvt \liff f(v)(a)=\tvt \right) \\
      &\iff \forall a\in\voc: ( v(a)=\tvt\ \liff ( \exists w\in\allv_2: f(w)(a)=\tvt \\
      & \hspace*{6.4cm} \land v=w ) ) \\
      &\iff \forall a\in\voc: ( v(a)=\tvt\ \liff ( \exists w\in\allv_2: f(w)(a)=\tvt \\
      & \hspace*{5.8cm} \land v(\phi_w)=\tvt ) ) \\
      &\iff \forall a\in\voc: \left( v(a)=\tvt\ \liff v\!\left(\mathop{\bigvee_{w\in\allv_2,}}_{f(w)(a)=\tvt}\phi_w\right)=\tvt \right) \\
      &\iff \forall a\in\voc: v(a)=v\!\left(\mathop{\bigvee_{w\in\allv_2,}}_{f(w)(a)=\tvt}\phi_w\right) \\
      &\iff \forall a\in\voc: v(a)=v(\varphi_a^f)
      \iff v\in\mod(D_f) \hspace*{7mm} \Box
    \end{align*}
  \end{longproof}
\end{proposition}
A related result was given by \citet[Proposition~10]{Strass2015}.
The characterization we presented here fits into the general framework of this paper and is directly usable for our realizability algorithm.
Wrapping up, the next result summarizes how ADF realizability can be captured by different types of characterizations for the semantics we considered so far.
\begin{theorem}\label{th:main}
  Let \mbox{$V\subseteq\allv$} be a set of interpretations and consider \mbox{$\sigma\in\{adm,com,mod\}$}.
  There is an ADF \adf such that \mbox{$\sigma(\adf)=V$} if and only if there is a $\sigma$-characterization for $V$.
\end{theorem}
The preferred semantics of an ADF \adf is closely related to its admissible semantics as,
by definition, the preferred interpretations of \adf are its \ileq-maximal admissible interpretations.
As a consequence we can also describe preferred realizability in terms of $\adm$-characterizations.
We use the lattice-theoretic standard notation $\topv{V}$ to select the \ileq-maximal
elements of a given set $V$ of interpretations.
\begin{corollary}\label{co:prefReal}
Let \mbox{$V\subseteq\allv$} be a set of interpretations.
There is an ADF \adf with \mbox{$\pre(\adf)=V$} iff there is an $\adm$-character\-ization for some \mbox{$V'\subseteq\allv$} with \mbox{$V\subseteq V'$} and \mbox{$\topv{V'}=V$}.
\end{corollary}
\iflong{
Finally, we give a result on the complexity of deciding realizability for the mentioned formalisms and semantics.

\begin{proposition}
  \label{thm:complexity:realizability}
  Let \mbox{$\F\in\set{\AF,\SETAF,\BADF,\ADF}$} be a formalism and
  \mbox{$\sigma\in\set{\adm,\com,\prf,\mod}$} be a semantics.
  The decision problem
  ``Given a vocabulary $\voc$ and a set \mbox{$V\subseteq\allv$} of interpretations over $\voc$, is there a \mbox{$\kb\in\F$} such that \mbox{$\sigma(\kb)=V$}?''
  can be decided in nondeterministic time that is polynomial in the size of $V$.\footnote{We assume here that the representation of any $V$ over $\voc$ has size $\Theta(3^{\abs{\voc}})$. There might be specific $V$ with smaller representations, but we cannot assume any better for the general case.}
  \begin{proof}
    For all considered $\F$ and $\sigma$, computing all $\sigma$-interpretations of a given witness \mbox{$\kb\in\F$} can be done in time that is linear in the size of $V$.
    Comparing the result to $V$ can also be done in linear time.
    \hfill$\Box$
  \end{proof}
\end{proposition}
}\fi

\subsection{Deciding Realizability: \Cref{alg:main}}
\label{sec:main-algorithm}

Our \iflong{ main }\fi algorithm for deciding realizability is a propagate-and-guess algorithm in the spirit of the DPLL algorithm for deciding propositional satisfiability~\citep{DBLP:reference/fai/GomesKSS08}.
It is generic with respect to (1) the formalism $\F$ and (2) the semantics $\sigma$ for which should be realized.
To this end, the propagation part of the algorithm is kept exchangeable and will vary depending on formalism and semantics.
Roughly, in the propagation step the algorithm uses the desired set $V$ of interpretations to derive certain necessary properties of the realizing knowledge base (line~\ref{alg:l:propagate}).
This is the essential part of the algorithm:
the derivation rules (\emph{propagators}) used there are based on characterizations of realizability with respect to formalism and semantics.
Once propagation of properties has reached a fixed point (line~\ref{alg:l:fixed}), the algorithm checks whether the derived information is sufficient to construct a knowledge base.
If so, \iflong{the knowledge base can be }\fi constructed and returned (line~\ref{alg:l:success}).
Otherwise (no more information can be obtained through propagation and there is not enough information to construct a knowledge base yet), the algorithm guesses another assignment for the characterization (line~\ref{alg:l:guess}) and calls itself recursively.

\begin{figure*}[t!]
  \centering
  {\smaller
    \begin{align*}
      p^{\in}_{adm}(V,F) &= \{(v_2,a,v(a)) \ \mid\ v\in V, v_2\in \tve{v},v(a)\neq\tvu\}                           &\hspace*{8mm}&     p^{\in,\tvu}_{\com}(V,F)=\{ (v_2,a,\neg \tvx)\ \mid\ v\in V, v_2\in \tve{v},v(a)=\tvu,\\                                       
      p^{\notin}_{adm}(V,F) &= \{(v_2,a,\neg v(a)) \ \mid\ v\in \allv\setminus V, v_2\in \tve{v},                   &&    \qquad \tvx\in\{\tvt,\tvf\},\forall v'_2\in \tve{v}: v_2\neq v'_2  \limplies (v'_2,a,\tvx)\in F\}\\                            
                         &\qquad  v(a)\neq \tvu,\forall b\in\voc\setminus v^{-1}(\tvu), \forall v'_2\in \tve{v}:  &&      p^{\not\in,\tvt\tvf}_{\com}(V,F)=\{(v_2,a,\neg v(a)) \ \mid\  v\in\allv\setminus V, v_2\in \tve{v}, v(a)\neq\tvu, \\           
                         &\qquad (a,v_2)\neq(b,v'_2) \limplies  (v'_2,b,v(b))\in F\}                             &&       \qquad \forall b\in \voc\setminus v^{-1}(\tvu), \forall v'_2\in \tve{v}: (a,v_2)\neq(b,v'_2) \limplies  (v'_2,b,v(b))\in F, \\ 
      p^\lightning_\adm(V, F) &= \{ (v,a,\tvt),(v,a,\tvf) \ \mid\ v\in\allv_2, a\in\voc, v_\tvu\not\in V \}          &&   \qquad \forall b\in v^{-1}(\tvu), \exists v''_2,v'''_2\in \tve{v} : (v''_2,b,\tvt),(v'''_2,b,\tvf)\in F \} \\                  
      p^{\in}_\mod(V, F) &= \{ (v, a, v(a)) \ \mid\ v\in V, a\in\voc \}                                            &&     p^{\not\in,\tvu}_{\com}(V,F)=\{(v_2,a,\neg \tvx)\ \mid\  v\in\allv\setminus V, v_2\in \tve{v}, v(a)=\tvu, \\                   
      p^{\notin}_\mod(V, F) &= \{ (v, a, \neg v(a)) \ \mid\ v\in\allv_2\setminus V, a\in\voc,                      &&    \qquad \forall b\in \voc\setminus v^{-1}(\tvu),\forall v'_2\in \tve{v}: (v_2,b,v(b))\in F,\\                                   
                         &\qquad\forall c\in\voc\setminus\set{a} : (v,c,v(c))\in F \}                           &&       \qquad \forall b\in v^{-1}(\tvu)\setminus\{a\}: \exists v''_2,v'''_2\in \tve{v} :(v''_2,b,\tvt),\\                             
      p^\lightning_\mod(V, F) &= \{ (v,a,\tvt),(v,a,\tvf) \ \mid\ v\in\allv_2, a\in\voc, V\not\subseteq\allv_2 \}  &&    \qquad (v'''_2,b,\tvf)\in F, \forall v''''_2\in \tve{v}\setminus\{v_2\}: (v''''_2,b,\tvx)\in F \} 
    \end{align*}
    \iflong{\vspace*{-2em}}\fi
  }
  \caption{Semantics propagators for the
    complete (\mbox{$\propagator{\ADF}{\com}=\{p^{\in,\tvt\tvf}_{\com},p^{\in,\tvu}_{\com},p^{\not\in,\tvt\tvf}_{\com},p^{\not\in,\tvu}_{\com}\}$} with \mbox{$p^{\in,\tvt\tvf}_{\com}(V,F)=p^{\in}_{\adm}(V,F)$}),
    admissible (\mbox{$\propagator{\ADF}{\adm}=\{p^{\in}_{\adm},p^{\notin}_{\adm},p^\lightning_{\adm}\}$}),
    and
    model semantics (\mbox{$\propagator{\ADF}{\mod} = \{ p^{\in}_\mod, p^{\notin}_\mod, p^\lightning_\mod \}$}).
  }
  \label{fig:semantics-propagators}
\end{figure*}%

The main data structure that \Cref{alg:main} operates on is a set of triples $(v, a, \tvx)$ consisting of a two-valued interpretation \mbox{$v\in\allv_2$}, an atom \mbox{$a\in\voc$} and a truth value \mbox{$\tvx\in\two$}.
This data structure is intended to represent the $\sigma$-characterizations introduced in \Cref{def:comChar,def:admChar,def:modChar}.
There, a $\sigma$-characterization is a function \mbox{$f:\allv_2\to\allv_2$} from two-valued interpretations to two-valued interpretations.
However, as the algorithm builds the $\sigma$-characterization step by step and there might not even be a $\sigma$-characterization in the end (because $V$ is not realizable), we use a set $F$ of triples $(v, a, \tvx)$ to be able to represent both partial and incoherent states of affairs.
The $\sigma$-characterization candidate induced by $F$ is partial if we have that for some $v$ and $a$, neither \mbox{$(v,a,\tvt)\in F$} nor \mbox{$(v,a,\tvf)\in F$};
likewise, the candidate is incoherent if for some $v$ and $a$, both \mbox{$(v,a,\tvt)\in F$} and \mbox{$(v,a,\tvf)\in F$}.
If $F$ is neither partial nor incoherent, it gives rise to a unique $\sigma$-characterization that can be used to construct the knowledge base realizing the desired set of interpretations.
The correspondence to the characterization-function is\iflong{ then such that }\fi\ \mbox{$f(v)(a) = \tvx$} iff \mbox{$(v,a,\tvx) \in F$}.

In our presentation of the algorithm we focused on its main features, therefore the guessing step (line~\ref{alg:l:guess}) is completely ``blind''.
It is possible to use common CSP techniques, such as shaving (removing guessing possibilities that directly lead to inconsistency).
Finally, we remark that the algorithm can be extended to enumerate all possible realizations of a given interpretation set -- by keeping all choice points in the guessing step and thus exhaustively exploring the whole search space.

\newfloat{algorithm}{t}{lop}
\renewcommand{\algorithmicrequire}{\textbf{Input:}}
\renewcommand{\algorithmicensure}{\textbf{Output:}}
\begin{algorithm}
  \caption{$\realize(\F,\sigma, V, F)$}
  \label{alg:main}
\iflong
  \begin{algorithmic}[1]
    \REQUIRE
    \begin{minipage}[t]{0.92\linewidth}
      \begin{compactitem}
      \item a formalism $\F$
      \item a semantics $\sigma$ for $\F$
      \item a set \mbox{$V$} of interpretations \mbox{$v:\voc\to\three$}
      \item a relation \mbox{$F\subseteq\allv_2\times\voc\times\two$}, initially empty
      \end{compactitem}
    \end{minipage}
    \ENSURE 
    \begin{minipage}[t]{0.9\linewidth}
      a $\kb\in\F$ with $\sigma(\kb)=V$ or ``no'' if none exists
    \end{minipage}
    \REPEAT 
    \STATE \label{alg:l:propagate} {\bf set} $F_\Delta := \bigcup\limits_{p\in\propagator{\F}{\sigma}} p(V,F)\setminus F$
    \STATE \label{alg:l:setF} {\bf set} $F := F\cup F_\Delta$
    \IF {$\exists v\in\allv_2,\exists a\in\voc: \set{(v,a,\tvt),(v,a,\tvf)}\subseteq F$} \STATE \label{alg:l:failure} {\bf return} {``no''} \ENDIF
    \UNTIL{ \vspace{-3pt} $F_\Delta=\emptyset$ \label{alg:l:fixed}}
    \IF {$\forall v\in\allv_2,\forall a\in\voc,\exists x\in\two:(v,a,x)\in F$}\label{alg:l:functional} \STATE \label{alg:l:success} \vspace{-6pt}{\bf return} $\canonKB{\F}{\sigma}(F)$ \ENDIF
    \vspace{-3pt} \STATE \label{alg:l:guess} {\bf choose}~\mbox{$v\in\allv_2,a\in\voc$}~with~\mbox{$(v,a,\tvt)\notin F$},~\mbox{$(v,a,\tvf)\notin F$}%
    \IF {$\realize(\F,\sigma,V, F\cup\set{(v,a,\tvt)})\neq \text{``no''}$}
        \STATE {\bf return} $\realize(\F,\sigma,V, F\cup\set{(v,a,\tvt)})$
    \ELSE
     \STATE {\bf return} $\realize(\F,\sigma,V, F\cup\set{(v,a,\tvf)})$
    \ENDIF
  \end{algorithmic}
\fi
\end{algorithm} 

In the case where the constructed relation $F$ becomes functional at some point, the algorithm returns a realizing knowledge base $\canonKB{\F}{\sigma}(F)$.
For ADFs, this just means that we denote by $f$ the $\sigma$-characterization represented by $F$ and set \mbox{$\canonKB{\ADF}{\sigma}(F)=D^f$}.
For the remaining formalisms we will introduce the respective constructions in later subsections.

\iflong{
The algorithm is parametric in two dimensions, namely with respect to the formalism $\F$ and with respect to the semantics $\sigma$.
These two aspects come into the algorithm via so-called \emph{propagators}.
A propagator is a formalism-specific or semantics-specific set of derivation rules.
}\fi
Given a set $V$ of desired interpretations and a partial $\sigma$-characterization $F$, a propagator $p$ derives new triples $(v,a,\tvx)$ that must necessarily be part of any total $\sigma$-characterization $f$ for $V$ such that $f$ extends $F$.
In the following, we present semantics propagators for admissible, complete and two-valued model (in (SET)AF terms stable) semantics, and formalism propagators for BADFs, AFs, and \SETAF{}s.%

\subsection{Semantics Propagators}
\label{sec:semantics-propagators}

These propagators (cf.\ Figure~\ref{fig:semantics-propagators})
are directly derived from the properties
of $\sigma$-characterizations presented in
\Cref{def:comChar,def:admChar,def:modChar}.
While the definitions provide exact conditions to check
whether a given function is a $\sigma$-characterization,
the propagators allow us to derive definite values of partial characterizations that are necessary to fulfill the conditions for being a $\sigma$-characterization.

For admissible semantics, the condition for a function
$f$ to be an $\adm$-characterization of a desired set of interpretations $V$
(cf.\ Definition~\ref{def:admChar})
can be split into a condition for desired interpretations \mbox{$v \in V$}
and two conditions for undesired interpretations \mbox{$v \notin V$}.
Propagator $p^{\in}_{adm}$ derives new triples
by considering interpretations $v \in V$.
Here, for all two-valued interpretations $v_2$ that extend $v$,
the value $f(v_2)$ has to be in accordance with $v$ on $v$'s Boolean part,
that is, the algorithm adds $(v_2,a,v(a))$ whenever $v(a) \neq \tvu$.
On the other hand, $p^{\notin}_{adm}$ derives new triples
for $v \notin V$ in order to ensure that there is a
two-valued interpretation $v_2$ extending $v$
where $f(v_2)$ differs from $v$ on a Boolean value of $v$.
Note that while $p^{\in}_{adm}$ immediately allows us to derive information about $F$
for each desired interpretation \mbox{$v \in V$}, propagator $p^{\notin}_{adm}$ is much weaker in the sense that it only derives a triple of $F$ if there is no other way to meet the conditions for an undesired interpretation. 
Special treatment is required for the interpretation $v_\tvu$ that maps all statements to $\tvu$ and is admissible for every ADF.
This is not captured by $p^{\in}_{adm}$ and $p^{\notin}_{adm}$ as these deal only with interpretations that have Boolean mappings.
Thus, propagator $p^\lightning_\adm$ serves to check whether \mbox{$v_\tvu\in V$}.
If this is not the case, the propagator immediately makes the relation $F$ incoherent and the algorithm correctly answers ``no''.

For complete semantics and interpretations \mbox{$v \in V$}, propagator $p^{\in,\tvt\tvf}_{\com}$ derives triples just like in the admissible case.
Propagator $p^{\in,\tvu}_{\com}$ deals with statements \mbox{$a \in \voc$} having \mbox{$v(a) = \tvu$} for which there have to be at least two \mbox{$v_2,v_2' \in \tve{v}$} having \mbox{$f(v_2)(a) = \tvt$} and \mbox{$f(v_2')(a) = \tvf$}.
Hence $p^{\in,\tvu}_{\com}$ derives triple $(v_2,a,\neg \tvx)$ if for all other \mbox{$v_2' \in \tve{v}$} we find a triple \mbox{$(v_2',a,\tvx)$}.
For interpretations \mbox{$v \notin V$} it must hold that there is some \mbox{$a \in \voc$} such that 
(i) \mbox{$v(a) \neq \tvu$} and \mbox{$f(v_2)(a) \neq v(a)$} for some \mbox{$v_2 \in \tve{v}$}
or
(ii) \mbox{$v(a) = \tvu$} but for all \mbox{$v_2 \in \tve{v}$}, $f(v_2)$ assigns the same Boolean truth value $\tvx$ to $a$.
Now if neither (i) nor (ii) can be fulfilled by any statement \mbox{$b \in \voc \setminus \{a\}$} due to the current contents of $F$,
propagators $p^{\not\in,\tvt\tvf}_{\com}$ and $p^{\not\in,\tvu}_{\com}$ derive triple
$(v_2,a,\neg v(a))$ for \mbox{$v(a) \neq \tvu$} if needed for $a$ to fulfill (i) and
$(v_2,a,\neg \tvx)$ for \mbox{$v(a) = \tvu$} if needed for $a$ to fulfill (ii), respectively. %
\begin{example}
Consider the set \mbox{$V_3=\{\tvu\tvu\tvu,\tvf\tvu\tvu,\tvu\tvu\tvf,\tvf\tvt\tvf\}$}.
First, we consider a run of $\realize(\ADF,\adm, V_3, \emptyset)$.
In the first iteration, propagator \mbox{$p^{\in}_{\adm}$} ensures that $F_\Delta$ in line~\ref{alg:l:propagate}
contains $(\tvf\tvf\tvf,a,\tvf)$, $(\tvf\tvt\tvf,a,\tvf)$, $(\tvf\tvt\tvf,c,\tvf)$, and $(\tvf\tvf\tvf,c,\tvf)$. 
Based on the latter three tuples and \mbox{$\tvf\tvu\tvf \notin V_3$}, propagator \mbox{$p^{\notin}_{\adm}$} derives $(\tvf\tvf\tvf,a,\tvt)$ in the second iteration
which together with $(\tvf\tvf\tvf,a,\tvf)$ causes the algorithm to return ``no''. Consequently, $V_3$ is not $\adm$-realizable.
A run of $\realize(\ADF,\com, V_3, \emptyset)$ on the other hand returns $\com$-characterization $f$ for $V_3$
that maps $\tvt\tvt\tvf$ to  $\tvt\tvf\tvf$, $\tvf\tvt\tvt$ to $\tvf\tvf\tvt$, $\tvf\tvt\tvf$ and $\tvf\tvf\tvf$ to $\tvf\tvt\tvf$ 
and all other \mbox{$v_2\in\allv_2$} to $\tvf\tvf\tvf$.
Hence, ADF $\adf_{f}$, given by the acceptance conditions
\iflong{
$$
\begin{array}{ll}
\varphi^f_a= a \land b \land \neg c,  \qquad  \varphi^f_c= \neg a \land b \land c,\\
\varphi^f_b= (\neg a \land b \neg \land \neg c) \lor (\neg a \land \neg b \land \neg c)\\
\end{array}
$$
}\fi
has $V_3$ as its complete semantics.
\end{example}
Finally, for two-valued model semantics,
propagator $p^{\in}_\mod$ derives new triples by looking at interpretations \mbox{$v\in V$}.
For those, we must find \mbox{$f(v)=v$} in each $\mod$-characterization $f$ by definition.
Thus the algorithm adds $(v,a,v(a))$ for each \mbox{$a\in\voc$} to the partial characterization $F$.
Propagator $p^{\notin}_\mod$ looks at interpretations \mbox{$v\in\allv_2\setminus V$}, for which it must hold that \mbox{$f(v)\neq v$}.
Thus there must be a statement \mbox{$a\in\voc$} with \mbox{$v(a)\neq f(v)(a)$}, which is exactly what this propagator derives whenever it is clear that there is only one statement candidate left.
This, in turn, is the case whenever all \mbox{$b\in\voc$} with the opposite truth value $\neg v(a)$ and all \mbox{$c\in \voc$} with \mbox{$c\neq a$} cannot coherently become the necessary witness any more.
The propagator $p^\lightning_\mod$ checks whether \mbox{$V\subseteq\allv_2$}, that is, the desired set of interpretations consists entirely of two-valued interpretations.
In that case this propagator makes the relation $F$ incoherent, following a similar strategy as $p^\lightning_\adm$.

\iflong{
\subsubsection{Preferred Semantics}

Realizing a given set of interpretations $V$ under preferred semantics requires special treatment.
We do not have a $\sigma$-characterization function for \mbox{$\sigma=\prf$} at hand
to directly check realizability of $V$ but
have to find some \mbox{$V' \subseteq \{v \in \allv \mid \exists v' \in V : v \ilt v'\}$}
such that \mbox{$V \cup V'$} is realizable under admissible semantics (cf.\ Corollary~\ref{co:prefReal}).
Algorithm~\ref{alg:prf2} implements this idea by
guessing such a $V'$ (line~\ref{alg:p2:guess}) 
and then using Algorithm~\ref{alg:main} to try to realize \mbox{$V \cup V'$}
under admissible semantics (line~\ref{alg:p2:call}).
If $\realize$ returns a knowledge base $\kb$ realizing \mbox{$V \cup V'$} under $\adm$
we can directly use $\kb$ as solution of $\realizePrf$
since it holds that \mbox{$\prf(\kb) = V$},
given that $V$ is an $\ileq$-antichain~(line~\ref{alg:p2:antichain}).

\begin{algorithm}
  \caption{$\realizePrf(\F, V)$}
  \label{alg:prf2}
  \begin{algorithmic}[1]
    \REQUIRE
    \begin{minipage}[t]{0.9\linewidth}
      \begin{itemize}
      \item a formalism $\F$
      \item a set \mbox{$V$} of interpretations \mbox{$v:\voc\to\three$}
	  \end{itemize}
    \end{minipage}
    \ENSURE 
    \begin{minipage}[t]{0.9\linewidth}
      Return some $\kb\in\F$ with $\prf(\kb)=V$ if one exists or ``no'' otherwise.
    \end{minipage}
    \IF {$\topv{V} \neq V$}
      \STATE \label{alg:p2:antichain} {\bf return} {``no''}
    \ENDIF
    \STATE {\bf set} $V^< := \{v \in \allv \mid \exists v' \in V : v \ilt v'\}$ %
    \STATE {\bf set} $X := \emptyset$
    \REPEAT
      \STATE \label{alg:p2:guess} {\bf choose}~\mbox{$V' \subseteq V^<$}~with~\mbox{$V' \notin X$}
      \STATE {\bf set} $X := X \cup \{V'\}$
      \STATE {\bf set} $V^\adm := V \cup V'$ %
      \IF {$\realize(\F,\adm,V^\adm,\emptyset) \neq \text{``no''}$}
        \STATE \label{alg:p2:call} {\bf return} $\realize(\F,\adm,V^\adm,\emptyset)$
      \ENDIF
    \UNTIL {$\forall V' \subseteq V^< : V' \in X$}
    \STATE {\bf return} {``no''}
  \end{algorithmic}
\end{algorithm}
}\fi

\begin{figure*}[t]
  \centering
  {\smaller
  \begin{align*}
    p^\SETAF(V, F) &= \set{ (v_\tvf, a, \tvt) \guard  a\in\voc } \cup
                     \set{ (w,a,\tvt) \guard  (v,a,\tvt)\in F, w\in\allv_2, w\tlt v } \cup
                     \set{ (w,a,\tvf) \guard (v,a,\tvf)\in F, w\in\allv_2, v\tlt w } \\
    p^\AF(V, F) &= p^\SETAF(V,F) \cup
                  \set{ (v_1\sqcup_t v_2,a,\tvt) \guard (v_1,a,\tvt)\in F, (v_2,a,\tvt)\in F } 
                  \qquad\qquad\ \ L^+ = \set{ (b,a) \guard (v,a,\tvf)\in F,v(b)=\tvf,(\update{v}{b}{\tvt},a,\tvt)\in F } \\
    p^\BADF(V,F) &= \{ (\update{v}{b}{\tvt},a,\tvx) \mid (v,a,\tvx)\in F, (w,a,\neg\tvx)\in F,
                   w(b)=\tvf, (\update{w}{b}{\tvt},a,\tvx)\in F \} 
                   \iflong\quad\!\!\fi\quad\, L^- = \set{ (b,a) \guard (v,a,\tvt)\in F,v(b)=\tvf,(\update{v}{b}{\tvt},a,\tvf)\in F }
  \end{align*}
  \iflong{\vspace*{-2em}}\fi
  }
  \caption{Formalism propagators. For formalism \mbox{$\F\in\set{\AF,\SETAF,\BADF}$} and any \mbox{$\sigma\in\set{\adm,\com,\prf,\mod}$}, we set the respective propagator for $\F$ to \mbox{$P^\F_\sigma = P^\ADF_\sigma \cup \set{p^\F}$} with $p^\F$ as defined above.
    $L^+$ and $L^-$ define link polarities for $\canonKB{\BADF}{\sigma}$.}
  \label{fig:formalism-propagators}
\end{figure*}
\subsection{Formalism Propagators}

When constructing an ADF realizing a given set $V$ of interpretations under a semantics $\sigma$,
the function $\canonKB{\ADF}{\sigma}(F)$ makes use of the $\sigma$-characterization given by $F$ in the following way:
$v$ is a model of the acceptance condition $\varphi_a$ if and only if we find \mbox{$(v,a,\tvt) \in F$}.
Now as bipolar ADFs, \SETAF{}s and AFs are all subclasses of ADFs by
restricting the acceptance conditions of statements,
these restrictions also carry over to the $\sigma$-characterizations.
The propagators defined below use structural knowledge on the form of acceptance conditions
of the respective formalisms to reduce the search space or to induce incoherence of $F$ whenever $V$ is not realizable.
\mysubsubsection{Bipolar ADFs}
For bipolar ADFs, we use the fact that each of their links must have at least one polarity, that is, must be supporting or attacking.
Therefore, if a link is not supporting, it must be attacking, and vice versa.
For canonical realization, we obtain the polarities of links,
i.e.\ the sets $L^+$ and $L^-$,
as defined in \Cref{fig:formalism-propagators}.
\mysubsubsection{AFs}
{\fussy To explain the AF propagators, we first need some more definitions.
On the two classical truth values, we define the truth ordering \mbox{$\tvf\tlt\tvt$}, whence the operations $\sqcup_t$ and $\sqcap_t$ with \mbox{$\tvf\sqcup_t\tvt = \tvt$} and \mbox{$\tvf\sqcap_t\tvt=\tvf$} result.
These operations can be lifted pointwise to two-valued interpretations as usual, that is,
\mbox{$(v_1\sqcup_t v_2)(a) = v_1(a)\sqcup_t v_2(a)$} and
\iflong{\mbox{$(v_1\sqcap_t v_2)(a) = v_1(a)\sqcap_t v_2(a)$}}\fi.
Again, the reflexive version of $\tlt$ is denoted by $\tleq$.
The pair $(\allv_2,\tleq)$ of two-valued interpretations ordered by the truth ordering forms a complete lattice with glb $\sqcap_t$ and lub $\sqcup_t$.}
\iflong{This complete lattice has the least element \mbox{$v_\tvf:\voc\to\set{\tvf}$}, the interpretation mapping all statements to false, and the greatest element \mbox{$v_\tvt:\voc\to\set{\tvt}$} mapping all statements to true, respectively.}\fi

Acceptance conditions of AF-based ADFs have the form of conjunctions of negative literals.
In the complete lattice $(\allv_2,\tleq)$, the model sets of AF acceptance conditions correspond to the lattice-theoretic concept of an \define{ideal}, a subset of $\allv_2$ that is downward-closed with respect to $\tleq$ and upward-closed with respect to $\sqcup_t$.
The propagator directly implements these closure properties:
application of $p^\AF$ ensures that when a $\sigma$-char\-ac\-teri\-za\-tion $F$
that is neither incoherent nor partial
is found in line~\ref{alg:l:functional} of Algorithm~\ref{alg:main},
then there is, for each \mbox{$a \in \voc$}, an interpretation $v_a$ such that
\mbox{$(v_a,a,\tvt) \in F$} and
\mbox{$v \tleq v_a$} for each \mbox{$(v,a,\tvt) \in F$}. %
Hence $v_a$ is crucial for the acceptance condition,
or in AF terms the attacks,
of $a$ and we can define
\mbox{$\canonKB{\AF}{\sigma}(F) = (\voc, \{(b,a) \mid a,b \in \voc, v_a(b) = \tvf\})$}.

\mysubsubsection{\SETAF{}s}
The propagator for \SETAF{}s, $p^\SETAF$, is a weaker version of that of AFs, since we cannot presume upward-closure with respect to $\sqcup_t$.
In \SETAF{}-based ADFs the acceptance formula is in \emph{conjunctive normal form}
containing only negative literals.
By a transformation preserving logical equivalence we obtain
an acceptance condition in \emph{disjunctive normal form}, again with only negative literals;
in other words, a \emph{disjunction} of AF acceptance formulas.
Thus, the model set of a \SETAF{} acceptance condition is not necessarily an ideal, but a union of ideals.
For the canonical realization we can make use of the fact
that, for each \mbox{$a \in \voc$}, the set
\mbox{$V^\tvt_a = \{v \in \allv_2 \mid (v,a,\tvt) \in F\}$}
is downward-closed with respect to $\tleq$,
hence the set of models of
\mbox{$\bigvee_{v \in \topvt{V^\tvt}} \bigwedge_{v(b)=\tvf} \neg b$}
is exactly $V^\tvt_a$.
The clauses of its corresponding CNF-formula exactly coincide with
the sets of arguments attacking $a$ in $\canonKB{\SETAF}{\sigma}(F)$.

\iflong{%
\subsection{Correctness}\fussy
For a lack of space, we could not include a formal proof of soundness and completeness of \Cref{alg:main}, but rather present arguments for termination and correctness.
\mysubsubsection{Termination}
With each recursive call, the set $F$ can never decrease in size, as the only changes to $F$ are adding the results of propagation in line~\ref{alg:l:setF} and adding the guesses in line~\ref{alg:l:guess}.
Also within the until-loop, the set $F$ can never decrease in size;
furthermore there is only an overall finite number of triples that can be added to $F$.
Thus at some point we must have \mbox{$F_\Delta=\emptyset$} and leave the until-loop.
Since $F$ always increases in size, at some point it must either become functional or incoherent, whence the algorithm terminates.
\mysubsubsection{Soundness}
If the algorithm returns a realizing knowledge base $\canonKB{\F}{\sigma}(F)$, then according to the condition in line~\ref{alg:l:functional} the relation $F$ induced a total function \mbox{$f:\allv_2\to\allv_2$}.
In particular, because the until-loop must have been run through at least once, there was at least one propagation step (line~\ref{alg:l:propagate}).
Since the propagators are defined such that they enforce everything that must hold in a $\sigma$-characterization, we conclude that the induced function $f$ indeed is a $\sigma$-characterization for $V$.
By construction, we consequently find that \mbox{$\sigma(\canonKB{\F}{\sigma}(F))=V$}.
\mysubsubsection{Completeness}
If the algorithm answers ``no'', then the execution reached line~\ref{alg:l:failure}.
Thus, for the constructed set $F$, there must have been an interpretation \mbox{$v\in\allv_2$} and a statement \mbox{$a\in\voc$} such that \mbox{$\set{(v,a,\tvt),(v,a,\tvf)}\subseteq F$}, that is, $F$ is incoherent.
Since $F$ is initially empty, the only way it could get incoherent is in the propagation step in line~\ref{alg:l:propagate}.
(The guessing step cannot create incoherence, since exactly one truth value is guessed for $v$ and $a$.)
However, the propagators are defined such that they infer only assignments (triples) that are necessary for the given $F$.
Consequently, the given interpretation set $V$ is such that either there is no realization within the ADF fragment corresponding to formalism $\F$ (that is, the formalism propagator derived the incoherence) or there is no $\sigma$-characterization for $V$ with respect to general ADFs (that is, the semantics propagator derived the incoherence).
In any case, $V$ is not $\sigma$-realizable for $\F$.%
}\fi

\iflong{
\section{Implementation}
\label{sec:implementation}

\begin{figure*}[t!]
\setlength{\columnsep}{1.3cm}
\begin{multicols}{2}

\begin{scriptsize}
\begin{Verbatim}[frame=topline,framesep=3mm,label=Main Encoding,labelposition=topline,numbers=left,samepage=false]
%
%
cterm(A, t(A)) :- s(A).
cterm(A, f(A)) :- s(A).

%
%
%
%
int(nil).
int((AS, I)) :- s(A), cterm(A, AS),
             int(I), smaller(A, I).
smaller(A, nil) :- s(A).
smaller(A, (H, I)) :- s(A), cterm(T, H), A < T, int(I).

%
member(T, (T, I)) :- int((T, I)).
member(T, (X, I)) :- int((X, I)), member(T, I).

%
int2(I) :- int(I), not hasU(I).
hasU(I) :- hasU(I, A).
hasU(I, A) :- int(I), s(A), not member(t(A), I),
                            not member(f(A), I).

%
ileq(I, J) :- int(I), int(J), not nileq(I, J).
nileq(I, J) :- int(I), int(J), member(T, I),
                           not member(T, J).

%
1 { ch(A, I, t); ch(A, I, f) } 1 :- s(A), int2(I).
\end{Verbatim}
\end{scriptsize}

\vfill\columnbreak

\begin{scriptsize}
\begin{Verbatim}[frame=topline,framesep=3mm,label=Two-Valued Model Encoding,labelposition=topline,numbers=left,samepage=false]
%
:- in(I), not int2(I).

%
ch(A, I, t) :- int2(I), in(I), s(A), member(t(A), I).
ch(A, I, f) :- int2(I), in(I), s(A), member(f(A), I).

%
ch(A, I, t) :- int2(I), not in(I), member(f(A), I), 
                ch(B, I, t) : s(B), member(t(B), I);
        ch(C, I, f) : s(C), member(f(C), I), C != A.
ch(A, I, f) :- int2(I), not in(I), member(t(A), I),
                ch(B, I, f) : s(B), member(f(B), I);
        ch(C, I, t) : s(C), member(t(C), I), C != A.
\end{Verbatim}
\end{scriptsize}

\begin{scriptsize}
\begin{Verbatim}[frame=topline,framesep=3mm,label=BADF Encoding,labelposition=topline,numbers=left,samepage=false]
%
%
att(B, A) :- ch(A, I, t), ch(A, J, f), diffFT(I, J, B).

%
%
sup(B, A) :- ch(A, I, f), ch(A, J, t), diffFT(I, J, B).

%
diffFT(I, J, A) :- int2(I), int2(J), member(f(A), I, B),
                                     member(t(A), J, B).
member(T, (T, I), I) :- int((T, I)).
member(T, (X, I), (X, B)) :- int((X, I)),
                 member(T, I, B), X != T.

%
%
ch(A, J, f) :- att(B, A), ch(A, I, f), diffFT(I, J, B).
ch(A, J, t) :- sup(B, A), ch(A, I, t), diffFT(I, J, B).
\end{Verbatim}
\end{scriptsize}

\end{multicols}
\setlength\columnsep{0.375in}
\caption{%
  Selected ASP encodings in clingo~4 syntax.
  The main encoding implements \Cref{alg:main}, the remaining encodings implement the two-valued model semantics propagator, and the BADF formalism propagator, respectively.
}\label{fig:encodings}
\end{figure*}

As Algorithm~\ref{alg:main} is based on propagation, guessing, and checking it is perfectly suited for an implementation using answer set programming (ASP)~\citep{N99,MT99}
as this allows for exploiting conflict learning strategies and heuristics of modern ASP solvers.
Thus, we developed ASP encodings in the \gringo language~\citep{gekakasc12a} for our approach.
Similar as the algorithm, our declarative encodings are modular, consisting of a main part responsible for constructing set $F$ and separate encodings for the individual propagators.
If one wants, e.g., to compute an AF realization under admissible semantics for a set $V$ of interpretations, an input program encoding $V$ is joined with the main encoding, the propagator encoding for admissible semantics as well as the propagator encoding for AFs.
Every answer set of such a program encodes a respective characterization function.
Our ASP encoding for preferred semantics is based on the admissible encoding and guesses further interpretations following the essential idea of Algorithm~\ref{alg:prf2}.
For constructing a knowledge base with the desired semantics, we also provide two ASP encodings that transform the output to an ADF in the syntax of the \diamondT tool~\citep{EllmauthalerS14}, respectively an AF in \aspartix syntax~\citep{EglyGW10,GagglMRWW15}.
Both argumentation tools are based on ASP themselves.
The encodings for all the semantics and formalisms we covered in the paper can be downloaded from \url{http://www.dbai.tuwien.ac.at/research/project/adf/unreal/}.
A selection of them is depicted in \Cref{fig:encodings} on the next page.
}\fi
\section{Expressiveness Results}
\label{sec:results}

In this section we briefly present some results that we have obtained using our implementation.
We first introduce some necessary notation to describe the relative expressiveness of knowledge representation formalisms~\citep{gogic95comparative,Strass2015}.
For formalisms $\F_1$ and $\F_2$ with semantics $\sigma_1$ and $\sigma_2$, we say that $\F_2$ under $\sigma_2$ is at least as expressive as $\F_1$ under $\sigma_1$ and write \mbox{$\F_1^{\sigma_1} \eleq \F_2^{\sigma_2}$} if and only if \mbox{$\signature{\F_1}{\sigma_1}\subseteq\signature{\F_2}{\sigma_2}$}, where \mbox{$\signature{\F}{\sigma} = \set{ \sigma(\kb) \guard \kb\in\F }$} is the \define{signature of $\F$ under $\sigma$}.
As usual, we define \mbox{$\F_1\elt\F_2$} iff \mbox{$F_1\eleq\F_2$} and \mbox{$F_2\not\eleq\F_1$}.

\iflong{%
We now start by considering the signatures of AFs, \SETAF{}s and (B)ADFs for the unary vocabulary \mbox{$\set{a}$}\/:
\begin{align*}
  \signature{\AF}{\adm} = \signature{\SETAF}{\adm} &= \set{ \set{\tvu}, \set{\tvu, \tvt} } \\
  \signature{\AF}{\com} = \signature{\SETAF}{\com} &= \set{ \set{\tvu}, \set{\tvt} } \\
  \signature{\AF}{\prf} = \signature{\SETAF}{\prf} &= \set{ \set{\tvu}, \set{\tvt} } \\
  \signature{\AF}{\mod} = \signature{\SETAF}{\mod} &= \set{ \emptyset, \set{\tvt} } \\
  \signature{\ADF}{\adm} = \signature{\BADF}{\adm} &= \signature{\AF}{\adm} \cup \set{ \set{ \tvu, \tvf }, \set{ \tvu, \tvt, \tvf } } \\
  \signature{\ADF}{\com} = \signature{\BADF}{\com} &= \signature{\AF}{\com} \cup \set{ \set{ \tvf }, \set{ \tvu, \tvt, \tvf } } \\
  \signature{\ADF}{\prf} = \signature{\BADF}{\prf} &= \signature{\AF}{\prf} \cup \set{ \set{ \tvf }, \set{ \tvt, \tvf } } \\
  \signature{\ADF}{\mod} = \signature{\BADF}{\mod} &= \signature{\AF}{\mod} \cup \set{ \set{ \tvf }, \set{ \tvt, \tvf } }
\end{align*}
}\fi

The following result shows that the expressiveness of the formalisms under consideration
is in line with the amount of restrictions they impose on acceptance formulas.

\begin{theorem}
  \label{thm:eleq:af-setaf-badf-adf}
  For any \mbox{$\sigma\in\set{\adm,\com,\prf,\mod}$}\/:
  \iflong{
    \begin{enumerate}
    \item \mbox{$\AF^{\sigma}\elt\SETAF^{\sigma}$}.
    \item \mbox{$\SETAF^{\sigma}\elt\BADF^{\sigma}$}.
    \item \mbox{$\BADF^{\sigma}\elt\ADF^{\sigma}$}.
    \end{enumerate}%
  }\fi
  \begin{longproof}
(1)
      \mbox{$\AF^\sigma\eleq\SETAF^\sigma$} is clear (by modeling individual attacks via singletons).
      For \mbox{$\SETAF^\sigma\not\eleq\AF^\sigma$} the witnessing model sets over vocabulary \mbox{$\voc=\set{a,b,c}$} are
$\set{ \tvu\tvu\tvu, \tvt\tvt\tvf, \tvt\tvf\tvt, \tvf\tvt\tvt } \in \signature{\SETAF}{\sigma} \setminus \signature{\AF}{\sigma}$ and
$\set{ \tvt\tvt\tvf, \tvt\tvf\tvt, \tvf\tvt\tvt } \in \signature{\SETAF}{\tau} \setminus \signature{\AF}{\tau}$
      with \mbox{$\sigma \in \{\adm,\com\}$} and \mbox{$\tau \in \{\prf,\mod\}$}.
      By each pair of arguments of $\voc$ being $\tvt$ in at least one model,
      a realizing AF cannot feature any attack,
      immediately giving rise to the model $\tvt\tvt\tvt$.
      The respective realizing \SETAF{} is given by the attack relation
      \mbox{$R = \set{(\{a,b\},c), (\{a,c\},b), (\{b,c\},a)}$}.

(2)
      It is clear that \mbox{$\SETAF^\sigma\eleq\BADF^\sigma$} holds (all parents are always attacking).
      For \mbox{$\BADF^\sigma\not\eleq\SETAF^\sigma$} the respective counterexamples can be read off the signatures above:
      for \mbox{$\sigma\in\set{\adm,\com}$} we find
      \mbox{$\set{ \tvu, \tvt, \tvf }\in\signature{\BADF}{\sigma}
        \setminus\signature{\SETAF}{\sigma}$}
      and for \mbox{$\tau\in\set{\prf,\mod}$} we find
      \mbox{$\set{ \tvt, \tvf }\in\signature{\BADF}{\tau}\setminus\signature{\SETAF}{\tau}$}.

      (3)
      For \mbox{$\sigma=\mod$} the result is known~\citep[Theorem~14]{Strass2015};
      for the remaining semantics the model sets witnessing \mbox{$\ADF^\sigma\not\eleq\BADF^\sigma$} over vocabulary \mbox{$\voc=\set{a,b}$} are
      \begin{align*}
        \set{ \tvu\tvu, \tvt\tvu, \tvt\tvt, \tvt\tvf, \tvf\tvu } &\in \signature{\ADF}{\adm} \setminus \signature{\BADF}{\adm} \\
        \set{ \tvu\tvu, \tvt\tvu, \tvt\tvt, \tvt\tvf, \tvf\tvu } &\in \signature{\ADF}{\com} \setminus \signature{\BADF}{\com} \\
        \set{ \tvt\tvt, \tvt\tvf, \tvf\tvu } &\in \signature{\ADF}{\prf} \setminus \signature{\BADF}{\prf}
      \end{align*}
      A witnessing ADF is given by \mbox{$\varphi_a=a$} and \mbox{$\varphi_b=a\liff b$}.
      \hfill$\Box$
  \end{longproof}
\end{theorem}

\Cref{thm:eleq:af-setaf-badf-adf} is concerned with the relative expressiveness of the formalisms under consideration, given a certain semantics.
Considering different semantics we find that for all formalisms the signatures become incomparable:

\begin{proposition}
\label{prop:incomp_signatures}
\mbox{$\F_1^{\sigma_1} \not\eleq \F_2^{\sigma_2}$} and
\mbox{$\F_2^{\sigma_2} \not\eleq \F_1^{\sigma_1}$}
for all formalisms \mbox{$\F_1,\F_2 \in \{\AF,\SETAF,\BADF,\ADF\}$} and all semantics
\mbox{$\sigma_1, \sigma_2 \in \{\adm, \com, \prf, \mod\}$} with
\mbox{$\sigma_1 \neq \sigma_2$}.
\end{proposition}
\begin{longproof}
First, the result for $\adm$ and $\com$
follows by
\mbox{$\{\tvu,\tvt\} \in \Sigma^\adm_\AF$}, but \mbox{$\{\tvu,\tvt\} \notin \Sigma^\com_\ADF$}
and
\mbox{$\{\tvt\} \in \Sigma^\com_\AF$}, but \mbox{$\{\tvt\} \notin \Sigma^\adm_\ADF$}.
Moreover,
taking into account that the set of preferred interpretations (resp.\ two-valued models)
always forms a $\ileq$-antichain while
the set of admissible (resp.\ complete) interpretations never does,
the result follows
for \mbox{$\sigma_1 \in \{\adm,\com\}$} and \mbox{$\sigma_2 \in \{\prf,\mod\}$}.
Finally, %
since a \mbox{$\kb \in \F$} may not have any two-valued models
and a preferred interpretation is not necessarily two-valued,
the result for $\prf$ and $\mod$ follows.
\hfill$\Box$
\end{longproof}

Disregarding the possibility of realizing the empty set of interpretations
under the two-valued model semantics,
we obtain the following relation for ADFs.

\begin{proposition}
\label{prop:mod_in_prf_adf}
$(\signature{\ADF}{\mod} \setminus \{\emptyset\}) \subseteq \signature{\ADF}{\prf}$.
\end{proposition}
\begin{longproof}
Consider some \mbox{$V \in \signature{\ADF}{\mod}$} with \mbox{$V \neq \emptyset$}.
Clearly \mbox{$V\subseteq\allv_2$} and by \Cref{prop:modReal} there is a $\mod$-characterization \mbox{$f:\allv_2\to\allv_2$} for $V$, that is, \mbox{$f(v) = v$} iff \mbox{$v \in V$}.
Define \mbox{$f':\allv_2 \to \allv_2$} such that \mbox{$f'(v)=f(v)=v$} for all \mbox{$v \in V$}
and \mbox{$f'(v)(a) = \neg v(a)$} for all \mbox{$v \in \allv \setminus V$} and \mbox{$a \in \voc$}.
Now it holds that $f'$ is an $\adm$-characterization of
\mbox{$V' = \{v \in \allv \mid \forall v_2 \in \tve{v} : v_2 \in V\} \cup \{v_\tvu\}$}.
Since \mbox{$\topv{V'} = V$} we get that the ADF $D$ with acceptance formula
$\varphi^{f'}_a$ for each \mbox{$a \in \voc$} has \mbox{$\prf(D)=V$} whence \mbox{$V \in \signature{\ADF}{\prf}$}.
\hfill$\Box$
\end{longproof}

\noindent In contrast, this relation does not hold for AFs,
which was shown for extension-based semantics
by~\citet{LinsbichlerSW15}\iflong{~(Theorem~5)}\fi 
\ and immediately follows for the three-valued case.

\section{Discussion}
\label{sec:discussion}

\fussy
We presented a framework for realizability in which AFs, \SETAF{}s, BADFs and general ADFs can be treated in a uniform way.
The centerpiece of our approach is an algorithm for deciding realizability of a given interpretation-set in a formalism under a semantics.
The algorithm makes use of so-called propagators, by which it can be adapted to the different formalisms and semantics.
\iflong{%
We also presented an implementation of our framework in answer set programming and several novel expressiveness results that we obtained using our implementation.
}\fi
\iflong{In related work, \citet{PolbergPhD2016} studies a wide range of abstract argumentation formalisms, in particular their relationship with ADFs.
  This can be the basis for including further formalisms into our realizability framework:
  all that remains to do is figuring out suitable ADF fragments and developing propagators for them, just like we did exemplarily for \citeauthor{NielsenP06}'s \SETAF{}s.
  For further future work, we could also streamline existing propagators such that they do not only derive absolutely necessary assignments, but also logically weaker conclusions, such as disjunctions of (non-)assignments.
}\fi

\smaller

\end{document}